\documentclass[10pt,twocolumn,letterpaper]{article}

\usepackage[pagenumbers]{cvpr} 

\usepackage{graphicx}
\usepackage{amsmath}
\usepackage{amssymb}
\usepackage{booktabs}
\usepackage{siunitx} 
\usepackage{soul}
\usepackage[T1]{fontenc} 

\usepackage[pagebackref,breaklinks,colorlinks]{hyperref}

\usepackage[capitalize]{cleveref}
\crefname{section}{Sec.}{Secs.}
\Crefname{section}{Section}{Sections}
\Crefname{table}{Table}{Tables}
\crefname{table}{Tab.}{Tabs.}

\def\cvprPaperID{N} 
\def\confName{XAI4CV Workshop at CVPR}
\def\confYear{2023}

\begin{document}

\title{Towards Evaluating Explanations of Vision Transformers for Medical Imaging}

\author{
Piotr Komorowski\\
University of Warsaw\\
{\tt\small p.m.komorowski@gmail.com}
\and
Hubert Baniecki\\
University of Warsaw\\
{\tt\small h.baniecki@uw.edu.pl}
\and
Przemysław Biecek\\
University of Warsaw\\
Warsaw University of Technology\\
{\tt\small p.biecek@uw.edu.pl}
}
\maketitle

\begin{abstract}
As deep learning models increasingly find applications in critical domains such as medical imaging, the need for transparent and trustworthy decision-making becomes paramount. Many explainability methods provide insights into how these models make predictions by attributing importance to input features. As Vision Transformer (ViT) becomes a promising alternative to convolutional neural networks for image classification, its interpretability remains an open research question. This paper investigates the performance of various interpretation methods on a ViT applied to classify chest X-ray images. We introduce the notion of evaluating faithfulness, sensitivity, and complexity of ViT explanations. The obtained results indicate that Layerwise relevance propagation for transformers outperforms Local interpretable model-agnostic explanations and Attention visualization, providing a more accurate and reliable representation of what a ViT has actually learned. Our findings provide insights into the applicability of ViT explanations in medical imaging and highlight the importance of using appropriate evaluation criteria for comparing them.
\end{abstract}

\section{Introduction}
\label{sec:intro}

Vision Transformer (ViT)~\cite{dosovitskiy2021transformers} is a novel class of deep learning models that use self-attention mechanism \cite{attn} to process images in a sequence of patches, rather than relying on convolutional operations. ViTs have achived impressive results on image classification tasks~\cite{chen2021crossvit, dosovitskiy2021transformers}, including medical imaging~\cite{vanTulder2021MultiviewAO, xViTCOS}. However, as deep neural networks are increasingly deployed in critical domains such as healthcare, it is valuable to understand what features they rely on~\cite{hryniewska2021checklist}. Explainable Artificial Intelligence (XAI) methods aim to provide such understanding by generating human-interpretable representations of model behavior~\cite{LRP,lime,holzinger2022xai}.

Several XAI methods have been proposed or adapted for ViTs~\cite{attn_flow, Chefer_2021_CVPR}, yet a rigorous and standardized evaluation of these methods in terms of their quality of explanations is still lacking. To address this issue, we propose utilizing two key criteria, namely faithfulness~\cite{Metrics_Bhatt} and sensitivity~\cite{Metrics_Yeh, Metrics_Bhatt}, to assess the quality of the explanations. Moreover, there is no clear consensus on how to best visualize the attention maps produced by ViTs, which can be ambiguous or even misleading~\cite{bibal2022attention}.

\begin{figure*}[t]
    \centering
    \includegraphics[width=0.9\textwidth]{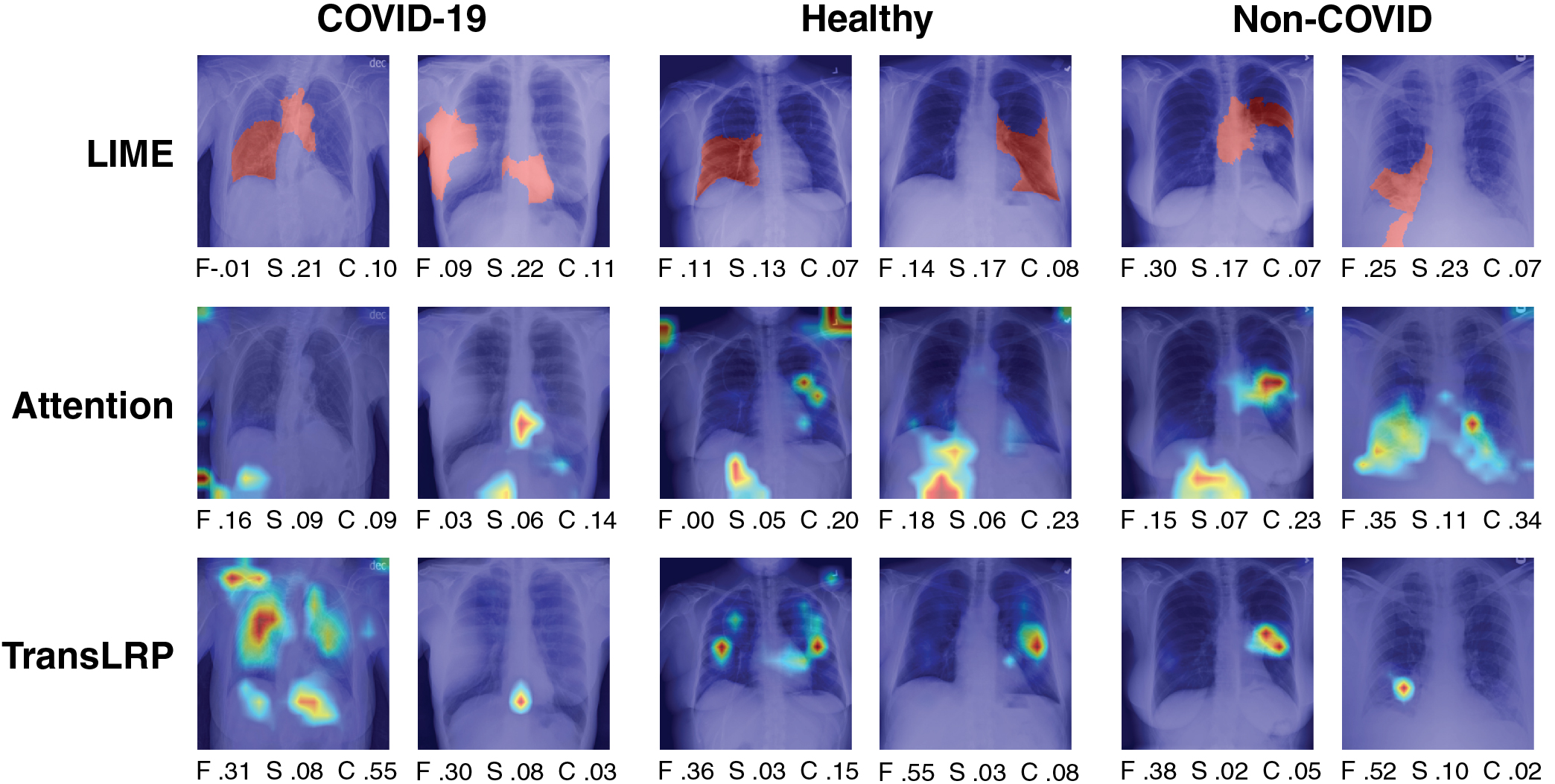}
    \caption{Exemplary explanations of ViT trained on the X-ray classification task. There are two images per class label (columns) and each one is explained using three interpretation methods (rows). For each explanation, we report their corresponding performance metrics measured with faithfulness (F), sensitivity (S), and complexity (C). For sensitivity and complexity lower scores are better, and for faithfulness higher ones. More such examples are presented in~\cref{app:explanations_supp}.}
    \label{fig:explanations}
\end{figure*}

This paper first introduces the notion of evaluating faithfulness, sensitivity, and complexity of ViT explanations. We then perform experiments with three interpretation methods: model-agnostic, attention-based and gradient-based applied to a ViT trained on a chest X-ray classification task (see examples in~\cref{fig:explanations}). Specifically, the obtained results show that a gradient-based approach outperforms baselines in both metrics providing more accurate and consistent explanations of ViT's decisions. Our goal is to foster the use of comprehensive metrics for a unified benchmarking protocol of ViT explanations, which is essential in medical imaging.

\section{Related work}
Prior to the emergence of ViTs, a variety of saliency map generation techniques were proposed for Convolutional Neural Networks (CNNs), including gradient-based~\cite{IG, SmoothGrad, FullGrad, GradCAM} and attribution propagation~\cite{DeepLIFT, DeepSHAP, CLRP, SGLRP} methods. Of the latter, Layerwise Relevance Propagation (LRP)~\cite{LRP} is a notable example that is formalized and theoretically justified in the Deep Taylor Decomposition framework~\cite{MONTAVON2017211}. Other methods such as Local Interpretable Model-agnostic Explanations (LIME) have been proposed as model-agnostic techniques for generating feature importance maps. Although fewer methods are available for ViTs, they exhibit interesting properties such as robustness against texture changes and severe occlusions compared to CNNs~\cite{Properties_Naseer}, making them a promising subject of XAI study.

The attention mechanism of ViTs can be easily visualized because its dimensions correspond to the input sequence dimension, making it a popular target for XAI adaptations. One of the most popular, the attention rollout method \cite{attn_flow} was introduced as a way to estimate precise attentions. Recently, a Layerwise Relevance Propagation for Transformers (TransLRP) approach~\cite{Chefer_2021_CVPR} was introduced to aggregate backward gradients and LRP throughout all layers and heads of the attention modules to derive explanation relevancy. This method outperforms previous transformer-specific and unspecific XAI methods and was extended to multimodal transformers~\cite{Chefer_2021_ICCV}. Although this approach is considered state-of-the-art in XAI for ViTs in various domains, many studies lack quantitative evaluation of the generated explanations~\cite{TransFER, Yang_2021_ICCV}. However, some studies have addressed this issue. For instance, Chefer~\etal utilized a perturbation metric that demonstrated the superiority of their method over others significantly~\cite{Chefer_2021_CVPR}.

In medical imaging, explanations often go unexamined. On Chest X-ray (CXR) images, a popular modality that our study focuses on, many studies were performed where generated explanations were used to explain predictions~\cite{xViTCOS, PARK2022102299}. However, the generated saliency maps were not evaluated and compared to objective baselines, which limits the reliability of the results. This lack of comprehensive evaluation also applies to computed tomography (CT)~\cite{xViTCOS, Dual_Decomp_Med_Images} and magnetic resonance imaging (MRI)~\cite{M3T, Dual_Decomp_Med_Images}. Existing evaluation methods either rely on segmentation masks~\cite{Benchmark_Saporta, Dual_Decomp_Med_Images}, which focus more on the model than the explanations, or use metrics such as the sensitivity of the model to weight randomization, repeatability, and reproducibility~\cite{Eval_Nishanth}. However, there remains a significant gap in the comprehensive evaluation of generated explanations, limiting the trustworthiness and applicability of these models in critical healthcare domains. We aim to address this gap in our study.

\section{Towards evaluating explanations of ViT}
We aim to perform a quantitative evaluation of the interpretation method adapted specifically to ViTs~\cite{Chefer_2021_CVPR}, in comparison to model-agnostic~\cite{lime} and attention-based interpretation methods \cite{bibal2022attention}, on the example of medical imaging.

\subsection{Explanation methods}
For the purpose of this study, we rely on three common approaches for generating feature attribution explanations.

As a well-proven baseline method, we use \textbf{LIME}~\cite{lime}, a model-agnostic approach that generates local, interpretable models to explain individual predictions. For image data, it selects superpixels using the k-lasso algorithm. In our study, we use binary saliency maps (see~\cref{fig:explanations}) and include the top two superpixels based on their positive contributions.

To visualize ViT \textbf{Attention}, we use the attention rollout technique proposed in~\cite{attn_flow}. At each ViT block, there is an attention matrix $A$ that describes how much attention flows from the previous layer to the next layer for each token. To obtain the total attention flow between two layers, these matrices can be multiplied. Since ViT utilizes residual connections, we model them by adding the identity matrix $I$ to the layer attention matrices $A+I$. To obtain one feature attribution matrix, it is suggested to average attention over multiple attention heads~\cite{attn_flow}, while other options like the minimum, maximum, or weighted average can also be considered. Following this, we recursively compute the attention rollout $\tilde{A}_l$ matrix at layer~$l$ as $\tilde{A}_l = (A_l + I) \tilde{A}_{l-1}$. The resulting matrix is of the shape $n\times n$, where $n$ represents a number of patches fed to the ViT. Since we focus on the classification task, only the classification token is considered. We take the row associated with this token, discard the first value, and reshape it to the patch grid size (for squared images, $\sqrt{n-1} \times \sqrt{n-1}$). Finally, we upsample it back to the size of the original image for visualization.

As a current state-of-the-art approach to explaining ViTs, we rely on the \textbf{TransLRP} algorithm proposed in~\cite{Chefer_2021_CVPR}. Attention scores are integrated throughout the attention graph, by combining relevancy and gradient information to iteratively eliminate negative contributions. The calculation of the output matrix $\bar{A_l}$ at layer $l$ can be expressed in a manner similar to previously with $\bar{A}_l = (A_l^{GR} + I) \bar{A}_{l-1}$ and $A_l^{GR} = (\nabla A_l \odot R_l)^+$ where $\odot$ denotes the Hadamard product, $\nabla A_l$ is the gradient of the attention map $A_l$, and $R_l$ represents the relevance of layer $l$ with respect to a target class. The notation $()^+$ is used to retain only positive values, which resemble positive relevance. To produce explanations, the final output is processed similarly to attention rollout. More details regarding the practical aspects of implementing these methods are provided in~\cref{app:exp_supp}.

\subsection{Explanation evaluation metrics}
Given feature attribution maps produced with various interpretation methods, we would like to evaluate their properties to assess their quality without human intervention. However, unlike supervised learning, where ground truth is available for comparison, this becomes an unsupervised problem. We formulate our evaluation framework based on three useful criteria: faithfulness to the model, sensitivity to the data, and relative complexity. In practice, we calculate these metrics by adapting their implementations conveniently available in the Quantus framework~\cite{hedstrom2023quantus}. 

\textbf{Faithfulness correlation} \cite{Metrics_Bhatt} aims to quantify the extent to which feature attributions accurately follow the model's prediction. It evaluates the linear correlation between the predicted logits of a modified test point when effectively removing features and the average explanation attribution for the particular subset of features, taking into account multiple runs and test samples. The metric generates a float value between $-1$ and $1$ for each input-attribution pair. \textbf{Average sensitivity} \cite{Metrics_Yeh, Metrics_Bhatt} ensures that if inputs are similar and their model outputs are close, their corresponding explanations should also be similar. To achieve this, Monte Carlo sampling-based approximation is employed while measuring how explanations change under minor perturbation of the input. \textbf{Effective complexity} \cite{Metrics_Nguyen} measures the number of attributions that exceed a specified threshold. Values above the threshold indicate that the corresponding features are important, while values below it suggest the features are not significant. Complexity is particularly important when visualization is the desired output of an explanation.

For class-specific methods like TransLRP and LIME, we generate explanations with respect to the model's predicted class. In contrast, the generated attributions do not differ across classes for Attention visualization. Therefore, we use an absolute value of the Faithfulness correlation metric in Attention visualization. As this technique is not tethered to any particular class, the unfavourable (negative) correlation could signify higher levels of faithfulness with respect to other classes. Our approach is an intuitive enhancement to the metric and may not be optimal for class-agnostic interpretation methods in general.

\section{Experiments with medical images}

We now critically evaluate ViT explanations for a medical imaging task, specifically chest X-ray classification.

\paragraph{Setup}

We use the \textbf{COVID-QU-Ex dataset}~\cite{Dataset_Tahir} consisting of $33{,}920$ CXR images including $11{,}956$ images with COVID-19 infection, $10{,}701$ healthy chest X-rays, and $11{,}263$ non-COVID infections, \ie, viral or bacterial pneumonia. The dataset authors' recommended $65/15/20$ split was adopted for training, validation, and test sets, respectively. We finetune a \textbf{base-sized ViT model}~\cite{dosovitskiy2021transformers} that was pre-trained on the ImageNet dataset~\cite{ImageNet}. The model achieved a test set accuracy of $0.954$, with balanced accuracy across all three classes. Details of model training are further described in~\cref{app:model_supp}. For calculating explanation evaluation metrics, $300$ images were randomly sampled from the test set, with $100$ images belonging to each class. 

\begin{table}
  \centering
  \caption{Evaluation results where lower sensitivity and higher faithfulness scores indicate better explanations. We report mean $\pm$ standard deviation over metric values calculated for $300$ images. TransLRP outperforms LIME and Attention visualization but provides more complex explanations.}
  \label{tab:main_results}
  \begin{tabular}{@{}lccc@{}}
    \toprule
    Explanation & Faithfulness$\uparrow$ & Sensitivity$\downarrow$ & Complexity$\downarrow$\\
    \midrule
    LIME & $0.07_{\pm 0.13}$ & $0.21_{\pm 0.06}$ & $0.09_{\pm 0.04}$\\
    Attention & $0.10_{\pm 0.08}$ & $0.07_{\pm 0.03}$ & $0.16_{\pm 0.12}$ \\
    TransLRP & $0.16_{\pm 0.18}$ & $0.06_{\pm 0.03}$ & $0.21_{\pm 0.16}$ \\
    \bottomrule
  \end{tabular}
\end{table}

\paragraph{Results}

\Cref{tab:main_results} presents a benchmark comparing interpretation methods in the three dimensions of interest. We observe that TransLRP outperforms the other techniques based in terms of two primary evaluation criteria: faithfulness, and to a lesser extent, sensitivity. This indicates that LRP provides feature attributions that are highly correlated with the model's predicted outcome, as reflected in its relatively high faithfulness score.  Moreover, as demonstrated by a low sensitivity metric value, its estimations remain the most consistent under small changes to input data.

Based on the \emph{quantitative results}, TransLRP appears to be a strong candidate for explaining ViT's predictions on the COVID-QU-Ex task, and thus can be recommended to be interpreted by humans, \eg, radiologist experts. A \emph{qualitative assessment} of explanation visualizations suggests that TransLRP is generally robust to imaging artifacts, although instances of spurious correlation leading to prediction can still occur (refer to \cref{fig:explanations}, column~1). LIME explanations are more consistent across images, but their reliance on superpixels may result in inaccuracies when the focus is not entirely on the lungs (see \cref{fig:explanations}, columns~2~\&~6). Attention visualization, when compared to TransLRP and LIME, appears to generate even more inaccurate explanations. Additional examples are available in~\cref{app:explanations_supp}.

\paragraph{On the quality-complexity tradeoff in interpretability}
In addition to the main metrics focusing on explanation quality, we also considered the complexity of the generated explanations, which becomes crucial when presenting visualizations to humans (see a discussion on information overload in~\cite{poursabzi2021manipulating}). \Cref{tab:main_results} reports that the Attention explanations are slightly less complicated than TransLRP. However, the complexity is quite low in both cases when compared to other aggregating methods of Attention heads (see \cref{tab:attention_heads}). For the purpose of this study, LIME attributions are binary, and therefore serve as a relatively simple baseline as indicated by the complexity metric. Note that the complexity metric's values depend significantly on a cutoff threshold and therefore will be biased toward binary feature attributions, which should be addressed in future work.

\paragraph{Are explanations equally good across classes?} Our study unveils the disparate quality of generated attributions across COVID-19, non-COVID, and healthy CXR cases as presented in~\cref{tab:class_results}. Plausibly, the overlapping symptomatic features of COVID pneumonia with other diseases based on medical imaging may limit the efficacy of generated attributions~\cite{hryniewska2021checklist}. Conversely, non-COVID pneumonia and healthy CXR cases with more unambiguous indications could facilitate the production of more accurate explanations. Such an analysis can be especially important to assess bias under an imbalanced class distribution. In each case, further explanation benchmarking is required to validate this challenge.

\begin{table}
  \centering
  \caption{Evaluation results for TransLRP divided into class labels. Explanations for the COVID-19 class perform visibly worst.}
  \label{tab:class_results}
  \begin{tabular}{@{}lccc@{}}
    \toprule
    Class & Faithfulness$\uparrow$ & Sensitivity$\downarrow$ & Complexity$\downarrow$\\
    \midrule
    COVID-19 & $0.07_{\pm0.13}$ & $0.07_{\pm0.03}$ & $0.34_{\pm0.19}$\\
    Healthy & $0.22_{\pm0.17}$  & $0.05_{\pm0.02}$ & $0.16_{\pm0.09}$ \\
    Non-COVID & $0.20_{\pm0.18}$ & $0.06_{\pm0.02}$ & $0.15_{\pm0.08}$ \\
    \bottomrule
  \end{tabular}
\end{table}

\begin{figure}
    \centering
    \includegraphics[width=0.9\linewidth]{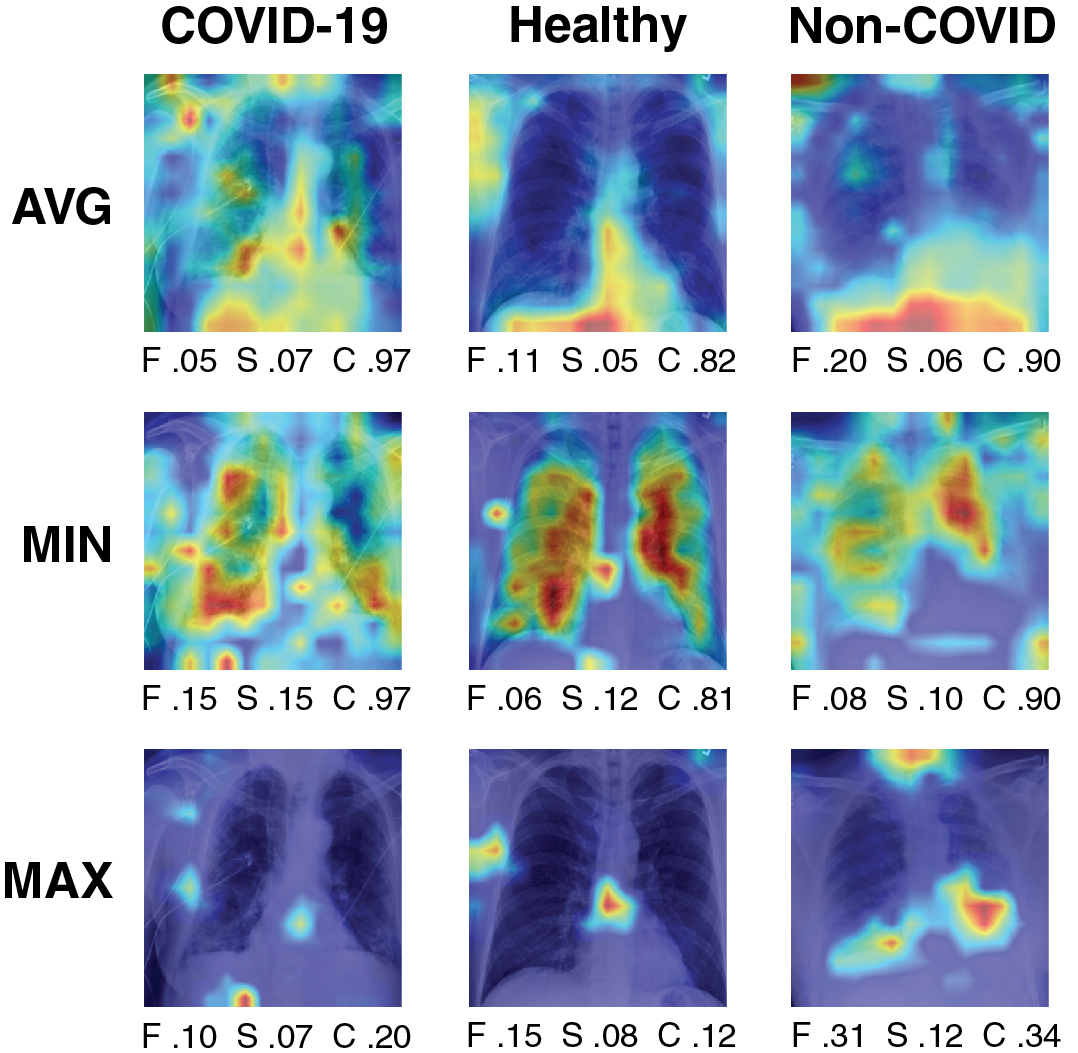}
    \caption{Exemplary explanations for Attention visualization divided into different approaches of its aggregation over ViT heads.}
    \label{fig:attention_heads}
\end{figure}

\paragraph{On the (im)proper attention aggregation over ViT heads}
The architecture of ViT utilizes multi-head self-attention and there are various methods to aggregate attention across the "heads" dimension. In~\cite{attn_flow}, it is proposed to average attention over ViT heads. However, our findings show that this approach produces overcomplex outcomes, as illustrated in \cref{fig:attention_heads}, where the entire image is highlighted. Even though minor input perturbations do not significantly impact the explanations, as shown by the sensitivity metric, their complexity renders them unsuitable. In~\cite{attn_vis}, attention rollout on natural images is analyzed, concluding that the best aggregation approach is to take the maximum over attention heads and discard a fraction of the lowest attributions. They also considered the minimum approach, which we included in our study. Ultimately, we have decided to adopt the maximum approach as it offers comparable explanation quality while maintaining significantly lower complexity (\cref{tab:attention_heads}).

\begin{table}
  \centering
  \caption{Evaluation results for Attention visualization divided into different approaches of its aggregation over ViT heads. Following~\cite{attn_vis}, we discard $99\%$ of the lowest attention values after aggregation with maximum, resulting in less complex explanations.}
  \label{tab:attention_heads}
  \begin{tabular}{@{}lccc@{}}
    \toprule
    Aggregation & Faithfulness$\uparrow$ & Sensitivity$\downarrow$ & Complexity$\downarrow$\\
    \midrule
    Average & $0.11_{\pm 0.09}$ & $0.06_{\pm 0.02}$ & $0.93_{\pm 0.06}$ \\
    Minimum & $0.11_{\pm 0.09}$ & $0.08_{\pm 0.04}$ & $0.63_{\pm 0.31}$ \\
    Maximum${-}$ & $0.10_{\pm 0.08}$ & $0.07_{\pm 0.03}$ & $0.16_{\pm 0.12}$ \\
    \bottomrule
  \end{tabular}
\end{table}

\section{Conclusion} Our study contributes to the advancement of the adoption of ViT explanations in practical fields, particularly in medicine where they are of great importance. The findings demonstrate that TransLRP outperforms other analyzed explanation methods on the COVID classification for CXR images. To our knowledge, this is the first work that comprehensively evaluates ViT explanations for medical imaging. In the future, this approach could be extended to other medical modalities, such as CT or MRI, for further exploration. The code is available at \url{https://github.com/piotr-komorowski/towards-evaluating-explanations-of-vit}.


\section*{Acknowledgment}
The work of Piotr Komorowski was supported by the European Union Regional Development Fund in the frame of the AI Tech project (POPC.03.02.00-00-0001/20-00) at the University of Warsaw, implemented under the supervision of the Chancellery of the Prime Minister of Poland. The experiments were conducted using the Entropy cluster funded by NVIDIA, Intel, the Polish National Science Center grant UMO2017/26/E/ST6/00622 and ERC Starting Grant TOTAL. This research was initiated as a team project with Szymon Antoniak and Kajetan Husiatyński during the XAI course at the University of Warsaw. We want to thank Maciej Chrabąszcz and Paulina Tomaszewska for their insightful comments.

{\small
\bibliographystyle{ieee_fullname}
\bibliography{references}

\begin{thebibliography}{10}\itemsep=-1pt

\bibitem{attn_flow}
Samira Abnar and Willem~H. Zuidema.
\newblock {Quantifying Attention Flow in Transformers}.
\newblock In {\em ACL}, 2020.

\bibitem{Eval_Nishanth}
Nishanth Arun, Nathan Gaw, Praveer Singh, Ken Chang, Mehak Aggarwal, Bryan
  Chen, Katharina Hoebel, Sharut Gupta, Jay Patel, Mishka Gidwani, Julius
  Adebayo, Matthew~D. Li, and Jayashree Kalpathy-Cramer.
\newblock {Assessing the Trustworthiness of Saliency Maps for Localizing
  Abnormalities in Medical Imaging}.
\newblock {\em Radiology: Artificial Intelligence}, 3(6):e200267, 2021.

\bibitem{Metrics_Bhatt}
Umang Bhatt, Adrian Weller, and José M.~F. Moura.
\newblock {Evaluating and Aggregating Feature-based Model Explanations}.
\newblock In {\em IJCAI}, 2020.

\bibitem{bibal2022attention}
Adrien Bibal, R{\'e}mi Cardon, David Alfter, Rodrigo Wilkens, Xiaoou Wang,
  Thomas Fran{\c{c}}ois, and Patrick Watrin.
\newblock {Is Attention Explanation? An Introduction to the Debate}.
\newblock In {\em ACL}, 2022.

\bibitem{Chefer_2021_ICCV}
Hila Chefer, Shir Gur, and Lior Wolf.
\newblock {Generic Attention-Model Explainability for Interpreting Bi-Modal and
  Encoder-Decoder Transformers}.
\newblock In {\em ICCV}, 2021.

\bibitem{Chefer_2021_CVPR}
Hila Chefer, Shir Gur, and Lior Wolf.
\newblock {Transformer Interpretability Beyond Attention Visualization}.
\newblock In {\em CVPR}, 2021.

\bibitem{chen2021crossvit}
Chun-Fu~(Richard) Chen, Quanfu Fan, and Rameswar Panda.
\newblock {CrossViT: Cross-Attention Multi-Scale Vision Transformer for Image
  Classification}.
\newblock In {\em International Conference on Computer Vision (ICCV)}, 2021.

\bibitem{ImageNet}
Jia Deng, Wei Dong, Richard Socher, Li-Jia Li, Kai Li, and Li Fei-Fei.
\newblock {ImageNet: A large-scale hierarchical image database}.
\newblock In {\em CVPR}, 2009.

\bibitem{dosovitskiy2021transformers}
Alexey Dosovitskiy, Lucas Beyer, Alexander Kolesnikov, Dirk Weissenborn,
  Xiaohua Zhai, Thomas Unterthiner, Mostafa Dehghani, Matthias Minderer, Georg
  Heigold, Sylvain Gelly, Jakob Uszkoreit, and Neil Houlsby.
\newblock {An Image is Worth 16x16 Words: Transformers for Image Recognition at
  Scale}.
\newblock In {\em ICLR}, 2021.

\bibitem{attn_vis}
Jacob Gildenblat.
\newblock Exploring explainability for vision transformers, 2020.
\newblock Accessed on March 12, 2023.

\bibitem{CLRP}
Jindong Gu, Yinchong Yang, and Volker Tresp.
\newblock {Understanding Individual Decisions of CNNs via Contrastive
  Backpropagation}.
\newblock In {\em ACCV}, 2018.

\bibitem{hedstrom2023quantus}
Anna Hedstr{\"{o}}m, Leander Weber, Daniel Krakowczyk, Dilyara Bareeva, Franz
  Motzkus, Wojciech Samek, Sebastian Lapuschkin, and Marina Marina~M.{-}C.
  H{\"{o}}hne.
\newblock {Quantus: An Explainable AI Toolkit for Responsible Evaluation of
  Neural Network Explanations and Beyond}.
\newblock {\em Journal of Machine Learning Research}, 24(34):1--11, 2023.

\bibitem{holzinger2022xai}
Andreas Holzinger, Anna Saranti, Christoph Molnar, Przemyslaw Biecek, and
  Wojciech Samek.
\newblock {\em Explainable AI Methods - A Brief Overview}, pages 13--38.
\newblock 2022.

\bibitem{hryniewska2021checklist}
Weronika Hryniewska, Przemysław Bombiński, Patryk Szatkowski, Paulina
  Tomaszewska, Artur Przelaskowski, and Przemysław Biecek.
\newblock {Checklist for responsible deep learning modeling of medical images
  based on COVID-19 detection studies}.
\newblock {\em Pattern Recognition}, 118:108035, 2021.

\bibitem{SGLRP}
Brian~Kenji Iwana, Ryohei Kuroki, and Seiichi Uchida.
\newblock {Explaining Convolutional Neural Networks using Softmax Gradient
  Layer-wise Relevance Propagation}.
\newblock In {\em ICCV Workshop}, 2019.

\bibitem{M3T}
Jinseong Jang and Dosik Hwang.
\newblock M3t: three-dimensional medical image classifier using multi-plane and
  multi-slice transformer.
\newblock In {\em CVPR}, 2022.

\bibitem{LRP}
Sebastian Lapuschkin, Alexander Binder, Grégoire Montavon, Frederick
  Klauschen, Klaus-Robert Müller, and Wojciech Samek.
\newblock {On Pixel-Wise Explanations for Non-Linear Classifier Decisions by
  Layer-Wise Relevance Propagation}.
\newblock {\em PLoS ONE}, 10:e0130140, 07 2015.

\bibitem{DeepSHAP}
Scott~M. Lundberg and Su{-}In Lee.
\newblock A unified approach to interpreting model predictions.
\newblock In {\em NeurIPS}, 2017.

\bibitem{xViTCOS}
Arnab~Kumar Mondal, Arnab Bhattacharjee, Parag Singla, and AP Prathosh.
\newblock {xViTCOS: Explainable Vision Transformer Based COVID-19 Screening
  Using Radiography}.
\newblock {\em IEEE Journal of Translational Engineering in Health and
  Medicine}, 10:1100110, 2022.

\bibitem{MONTAVON2017211}
Grégoire Montavon, Sebastian Lapuschkin, Alexander Binder, Wojciech Samek, and
  Klaus-Robert Müller.
\newblock Explaining nonlinear classification decisions with deep taylor
  decomposition.
\newblock {\em Pattern Recognition}, 65:211--222, 2017.

\bibitem{Properties_Naseer}
Muzammal Naseer, Kanchana Ranasinghe, Salman~H. Khan, Munawar Hayat,
  Fahad~Shahbaz Khan, and Ming{-}Hsuan Yang.
\newblock {Intriguing Properties of Vision Transformers}.
\newblock In {\em NeurIPS}, 2021.

\bibitem{Metrics_Nguyen}
An{-}phi Nguyen and Mar{\'{\i}}a~Rodr{\'{\i}}guez Mart{\'{\i}}nez.
\newblock On quantitative aspects of model interpretability.
\newblock {\em arXiv preprint arXiv:2007.07584}, 2020.

\bibitem{PARK2022102299}
Sangjoon Park, Gwanghyun Kim, Yujin Oh, Joon~Beom Seo, Sang~Min Lee, Jin~Hwan
  Kim, Sungjun Moon, Jae-Kwang Lim, and Jong~Chul Ye.
\newblock {Multi-task vision transformer using low-level chest X-ray feature
  corpus for COVID-19 diagnosis and severity quantification}.
\newblock {\em Medical Image Analysis}, 75:102299, 2022.

\bibitem{poursabzi2021manipulating}
Forough Poursabzi-Sangdeh, Daniel~G. Goldstein, Jake~M. Hofman,
  Jennifer~Wortman Wortman~Vaughan, and Hanna Wallach.
\newblock {Manipulating and Measuring Model Interpretability}.
\newblock In {\em CHI}, 2021.

\bibitem{lime}
Marco~Tulio Ribeiro, Sameer Singh, and Carlos Guestrin.
\newblock {``Why Should I Trust You?'': Explaining the Predictions of Any
  Classifier}.
\newblock In {\em KDD}, 2016.

\bibitem{Dual_Decomp_Med_Images}
Tom Ron, Michal Weiler-Sagie, and Tamir Hazan.
\newblock {Dual Decomposition of Convex Optimization Layers for Consistent
  Attention in Medical Images}.
\newblock In {\em ICML}, 2022.

\bibitem{Benchmark_Saporta}
Adriel Saporta, Xiaotong Gui, Ashwin Agrawal, Anuj Pareek, Steven~QH Truong,
  Chanh~DT Nguyen, Van-Doan Ngo, Jayne Seekins, Francis~G Blankenberg, Andrew~Y
  Ng, et~al.
\newblock {Benchmarking saliency methods for chest X-ray interpretation}.
\newblock {\em Nature Machine Intelligence}, pages 1--12, 2022.

\bibitem{GradCAM}
Ramprasaath~R. Selvaraju, Abhishek Das, Ramakrishna Vedantam, Michael Cogswell,
  Devi Parikh, and Dhruv Batra.
\newblock {Grad-CAM: Why did you say that? Visual Explanations from Deep
  Networks via Gradient-based Localization}.
\newblock In {\em ICCV}, 2017.

\bibitem{DeepLIFT}
Avanti Shrikumar, Peyton Greenside, and Anshul Kundaje.
\newblock Learning important features through propagating activation
  differences.
\newblock In {\em ICML}, 2017.

\bibitem{SmoothGrad}
Daniel Smilkov, Nikhil Thorat, Been Kim, Fernanda~B. Vi{\'{e}}gas, and Martin
  Wattenberg.
\newblock {SmoothGrad: removing noise by adding noise}.
\newblock In {\em ICML Workshop}, 2017.

\bibitem{FullGrad}
Suraj Srinivas and Fran{\c{c}}ois Fleuret.
\newblock Full-jacobian representation of neural networks.
\newblock In {\em NeurIPS}, 2019.

\bibitem{IG}
Mukund Sundararajan, Ankur Taly, and Qiqi Yan.
\newblock {Axiomatic Attribution for Deep Networks}.
\newblock In {\em ICML}, 2017.

\bibitem{Dataset_Tahir}
Anas~M. Tahir, Muhammad~E.H. Chowdhury, Amith Khandakar, Tawsifur Rahman, Yazan
  Qiblawey, Uzair Khurshid, Serkan Kiranyaz, Nabil Ibtehaz, M.~Sohel Rahman,
  Somaya Al-Maadeed, Sakib Mahmud, Maymouna Ezeddin, Khaled Hameed, and Tahir
  Hamid.
\newblock {COVID-19 infection localization and severity grading from chest
  X-ray images}.
\newblock {\em Computers in Biology and Medicine}, 139:105002, 2021.

\bibitem{vanTulder2021MultiviewAO}
Gijs van Tulder, Yao Tong, and Elena Marchiori.
\newblock Multi-view analysis of unregistered medical images using cross-view
  transformers.
\newblock In {\em ICCV}, 2021.

\bibitem{attn}
Ashish Vaswani, Noam Shazeer, Niki Parmar, Jakob Uszkoreit, Llion Jones,
  Aidan~N Gomez, \L~ukasz Kaiser, and Illia Polosukhin.
\newblock Attention is all you need.
\newblock In {\em NeurIPS}, 2017.

\bibitem{TransFER}
Fanglei Xue, Qiangchang Wang, and Guodong Guo.
\newblock Transfer: Learning relation-aware facial expression representations
  with transformers.
\newblock {\em CoRR}, abs/2108.11116, 2021.

\bibitem{Yang_2021_ICCV}
Sen Yang, Zhibin Quan, Mu Nie, and Wankou Yang.
\newblock {TransPose: Keypoint Localization via Transformer}.
\newblock In {\em ICCV}, 2021.

\bibitem{Metrics_Yeh}
Chih-Kuan Yeh, Cheng-Yu Hsieh, Arun Suggala, David~I Inouye, and Pradeep~K
  Ravikumar.
\newblock {On the (In)fidelity and Sensitivity of Explanations}.
\newblock In {\em NeurIPS}, 2019.

\end{thebibliography}
}

\clearpage
\appendix

\section{Supplementary explanations}\label{app:explanations_supp}

\begin{figure*}[!ht]
    \centering
    \includegraphics[width=0.9\textwidth]{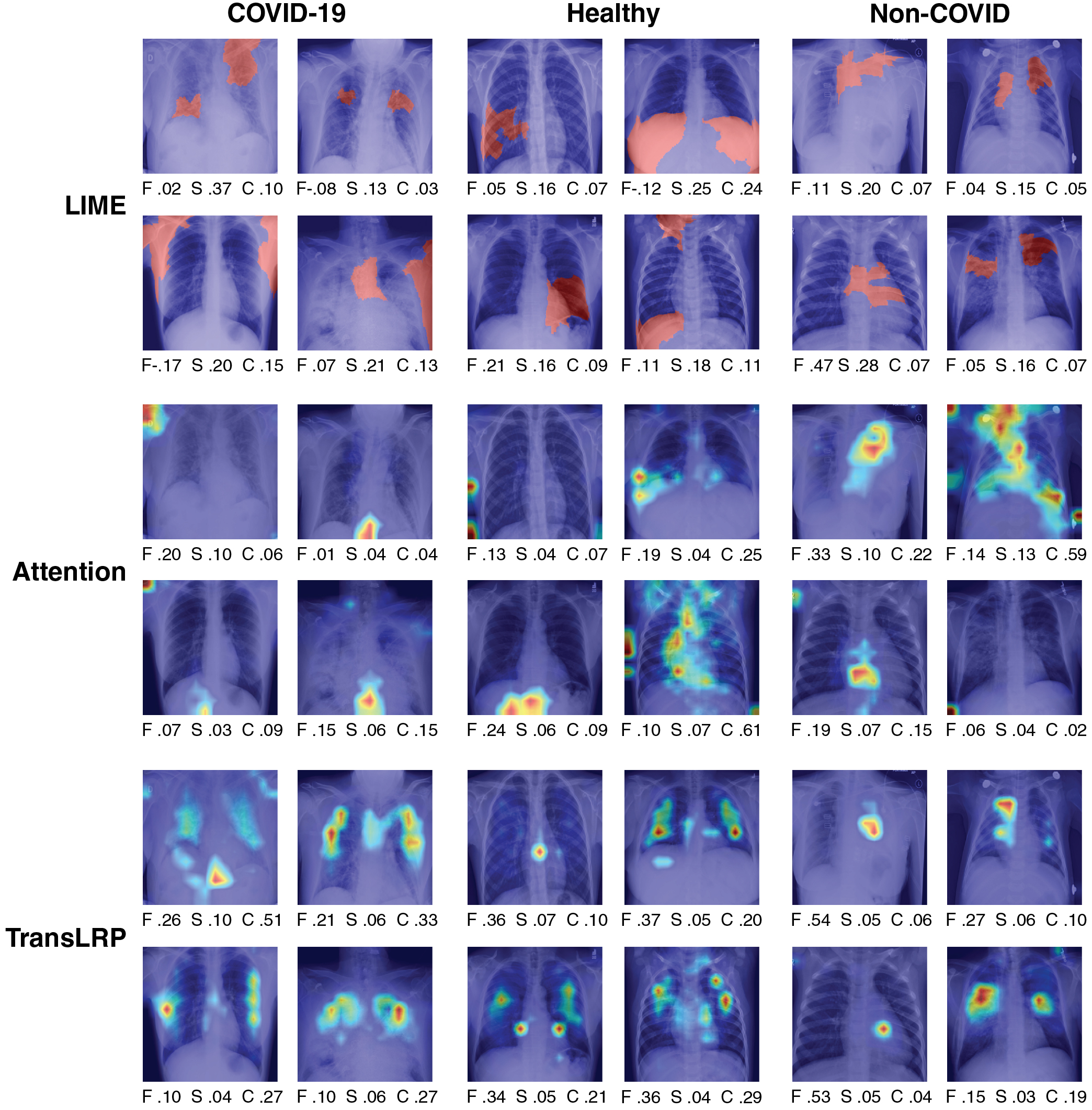}
    \caption{Exemplary explanations of ViT trained on the X-ray classification task. There are four images per class label (doubled columns) and each one is explained using three interpretation methods (doubled rows). For each explanation, we report their corresponding performance metrics measured with faithfulness (F), sensitivity (S), and complexity (C).}
    \label{fig:explanations_supp}
\end{figure*}

See \cref{fig:explanations_supp}.

\section{Implementation details of methods}\label{app:exp_supp}

In our study, we employed the original implementation of LIME \cite{lime} with $500$ samples generated per instance. We use binary saliency maps and include the top two superpixels based on their positive contributions. For attention visualization, we built on the implementation presented in \cite{attn_vis}. Our analysis involved three different aggregation approaches for ViT heads, namely average, minimum, and maximum. Following \cite{attn_vis}, in the case of maximum approach, after aggregation we discard $99\%$ of the lowest attention values, to enhance the isolation of salient regions in the image. For TransLRP we utilize the implementation from the original work \cite{Chefer_2021_CVPR}. Generated attributions for all methods were normalized to the $[0{,}1]$ range. All displayed examples of explanations are for input images where ViT accurately predicted the ground truth class label.

\section{Details of ViT training}\label{app:model_supp}

We use a base-sized ViT model that was pre-trained on ImageNet. The input consisted of a sequence of non-overlapping patches of size $16\times 16$ of the input image, followed by flattening and linear layers to produce a sequence of vectors. An additional token was added at the beginning of the sequence for classification. We fine-tuned the model for $15$ epochs with a batch size of $32$ and a constant learning rate of $3 \cdot 10^{-4}$. To prevent overfitting, we employed early stopping and found the best-performing model based on the validation set accuracy after the $10$th epoch of training. The model achieved an accuracy of $0.954$ on the test set with balanced accuracy among the three classes. To improve the model's generalization performance, we introduced light data augmentation techniques such as random crop and random rotation ($\pm15$ degrees), which effectively allowed the model to generalize better and achieve the best performance.

\end{document}